\documentclass[letterpaper, 10 pt, conference]{ieeeconf}  

\IEEEoverridecommandlockouts                              

\usepackage{graphics} 
\usepackage{epsfig} 
\usepackage{mathptmx} 
\usepackage{times} 
\usepackage{amsmath} 
\usepackage{amssymb}  
\usepackage{booktabs}

\usepackage[colorinlistoftodos,prependcaption,textsize=tiny]{todonotes}

\title{\LARGE \bf
What Am I? Evaluating the Effect of Language Fluency and Task Competency on the Perception of a Social Robot
}

\author{Shahira Ali, Haley N. Green, and Tariq Iqbal
\thanks{The authors are with the School of Engineering and Applied Science, University of Virginia, Charlottesville, VA, USA.
        {\tt\small \{sma6zw,hng9vf,tiqbal\}@virginia.edu}.}%
}

\usepackage{graphicx}
\usepackage{cite}
\usepackage{flushend} 

\begin{document}

\maketitle
\thispagestyle{empty}
\pagestyle{empty}

\begin{abstract}

Recent advancements in robot capabilities have enabled them to interact with people in various human-social environments (HSEs). In many of these environments, the perception of the robot often depends on its capabilities, e.g., task competency, language fluency, etc. To enable fluent human-robot interaction (HRI) in HSEs, it is crucial to understand the impact of these capabilities on the perception of the robot. Although many works have investigated the effects of various robot capabilities on the robot's perception separately, in this paper, we present a large-scale HRI study ($n = 60$) to investigate the combined impact of both language fluency and task competency on the perception of a robot. The results suggest that while language fluency may play a more significant role than task competency in the perception of the verbal competency of a robot, both language fluency and task competency contribute to the perception of the intelligence and reliability of the robot. The results also indicate that task competency may play a more significant role than language fluency in the perception of meeting expectations and being a good teammate. The findings of this study highlight the relationship between language fluency and task competency in the context of social HRI and will enable the development of more intelligent robots in the future.

\end{abstract}

\section{Introduction}

The integration of robots to interact and work alongside people has become a growing area of research \cite{bahishti,  yasar2020RAL, islam2021multigat, unhelkar2020semi, maven, caesar, kanda}. Advancements in robotics over the last few decades have enabled them to perform complex tasks and interact with people across various human-social environments (HSEs) \cite{eqamx, vader, graaf, imprint, patron, yasar2022HRI}. As such, there has been an increase in research understanding how robots are perceived in various scenarios based on their roles \cite{yan, Iqbal2016T-RO, iqbal_2021, anzalone, leite, friederike, posetron}.


One particular capability of robots that has widely been explored in the context of social human-robot interaction (HRI) is their verbal communication skills \cite{nikolaidis, vargas, woo,  haley_humor_aaai, zhao, haley_humor}. Along this research direction, many works have focused on investigating the communication abilities of robots, and their effectiveness in various task settings \cite{kanda, anzalone, vargas, breazeal}. Another work evaluated interpersonal communication and relationship building between a robot and a human partner and found that verbal communication capabilities had a positive impact on the social perception of the robot as well as can strengthen the bond between the collaborative pair \cite{breazeal}. Similarly, another study found that verbal communication positively impacts perceptions of friendliness and social presence in robots \cite{grigore}. Given the critical role verbal communication plays in HRI, many works in HRI have focused on making communication as natural and fluent as possible \cite{woo, vargas}. 

In human-human interactions, among various aspects of verbal communication, language fluency particularly plays a significant role in building a perception of a person. For example, some studies indicate that participants hold negative biases and stereotypes against non-native speakers \cite{gluszek, pavlenko, kang}. Although the impact of language fluency on the perception of a person in human-human interaction has been considerably studied, there has not been any work exploring the impact of language fluency on the perception of the robot. As robots are now expected to interact with people in various social scenarios, it's crucial to explore how verbal communication, specifically language fluency, may play a role in HRI.

\begin{figure}[t]
    \centering
    \includegraphics[width=0.45\textwidth]{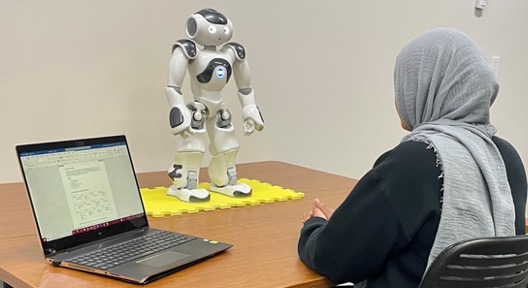}
    \caption{A NAO robot plays the ``What Am I?'' game with a human. The participant sits opposite the robot during the game and guesses the identity of the animal that the robot is assuming from the list provided to them.}
    \label{fig:setup}
    \vspace{-0.2in}
\end{figure}


Along with verbal communication being a significant part of interactions, robots are also expected to interact with people and perform various tasks with them. In these scenarios, the task competency of the robot also affects how people perceive the robot. Several studies have investigated varying task competency and its influence on the perception of robots \cite{van, walker, haring, kim, jung, waveren, carter}. For example, robots that were interruptive during tasks were perceived as being less task competent than the robots that completed the tasks without interleaving \cite{carter, kim}. Additionally, results from some works indicate how human agents perceived an erroneous robot as less trustworthy and less reliable than a competent robot in a collaborative setting \cite{salem}. 

Although many works focus on investigating the impact of a robot's verbal communication ability and task competency on the perception of the robot separately, to the best of our knowledge, no research to date has studied the combined impact of language fluency and task competency on the perception of a social robot. In this work, we are interested in exploring how varying language fluency and task competency in a social robot will affect people's perceptions of the robot. Specifically, we aim to investigate the impact of the robot's language fluency and task competency (i.e., fluent-competent, fluent-incompetent, disfluent-competent, and disfluent-incompetent conditions) on the perceptions of the robot's verbal competence, intelligence, and reliability along with whether the robot meeting expectations, being a good teammate, and if the participant is willing to work with the robot again. To investigate that, we have designed a human-robot interaction scenario modeled after the children's guessing game, ``What Am I?'', and conducted a large-scale HRI study ($n = 60$). In this task, a NAO robot describes various characteristics of an animal, and the human participant is prompted to guess what animal the robot is referring to. We vary the language fluency of the robot by allowing it to speak at a native or a non-native level and the task competency by allowing the robot to correctly match or fail to identify the animal. 


The results from the study suggest that while language fluency may play a greater role than task competency in the perception of verbal competence of a robot, both language fluency and task competency contribute to the perception of intelligence and reliability of the robot. The results also indicate that task competency may play a greater role than language fluency in the perception of meeting expectations and being a good teammate. These findings will allow us to have a deeper understanding of the relationship between language fluency and task competency on the perception of a robot and to develop ways to reduce non-native language biases and promote greater understanding and respect for linguistic diversity in the context of HRI.

\section{BACKGROUND}

\subsection{Language Fluency}\label{Language Fluency}

The definition of fluency in the context of spoken language in human-human interactions (HHI) has long been debated and studied. For example, Lennon \cite{lennon} differentiates between two approaches of fluency. The first is a holistic, broad sense in which fluency refers to global oral proficiency. The second portrays fluency in a much narrower sense and relies on the efficient and effortless planning and production of speech, such as speaking speed or grammatical accuracy. Furthermore, language fluency has been linked to higher levels of perceived proficiency and is ``best conceived of as fast, smooth, and accurate performance'' \cite{vanpatten, kormos}. However, there is a greater risk of confounding measures when multiple measures are used to define and examine fluency \cite{bosker}. Given the limited literature on definitive descriptions and classifications of language fluency in HHI, it is not surprising that language fluency in HRI has yet to be adequately explored. In this study, we will be referring to language fluency in Lennon's narrow sense and focus only on one measure of oral proficiency: \textit{grammatical accuracy}. 

\subsection{Language Bias}\label{Language Bias}

The study of language bias and perception has become increasingly important as globalization has led to a more diverse and multicultural society. One area of interest is the perception of non-native language speakers by native language speakers. For example, several studies have been conducted to investigate the extent of bias, and the factors that influence perceptions of non-native speakers \cite{gluszek, white, pavlenko, bongaerts, hagi, barona, munro}. Other studies depict how students express a preference for native English-speaking teachers and negatively judge the teaching skills of non-native English teachers \cite{rubin, todd, kang}. Another study found non-native speakers to have lower perceived intelligence and trustworthiness by native speakers \cite{james}. There is adequate literature surrounding native speaker bias and perceptions of non-native speakers in human-human interactions. However, little research has been done surrounding the impact of a person's linguistic abilities on their perceptions of a robot. A native speaker is someone who speaks that language with native fluency \cite{groff}. To reduce the bias of participants' linguistic background, in this work, we required all participants to be monolingual in English.  

\subsection{Language in Robots}\label{Language in Robots}

Verbal communication has been a crucial area of research in robotics, as it enables robots to interact with humans in a more natural and intuitive way \cite{nikolaidis, vargas, woo, zhao}. 
Language plays an important role in human-robot interaction, as it allows robots to understand and respond to human commands, convey information, and establish social relationships with humans \cite{jackson, scheutz_cantrell, bisk}. The focus of several works has been to improve perceptions of trust in robots by employing deep learning speech and automatically generated explanations \cite{wang, wenger}. With the importance of verbal communication and natural language established in HRI, it is important to understand how varying levels of language fluencies can impact perceptions of that robot. 

\subsection{Task Competency}\label{Task Competency}
Previous research in HRI has explored various aspects of robot behavior, including task performance, and their impact on the perception of the robot \cite{haley_humor,haley_humor_aaai}. For example, Salem et al. found that participants rated a robot as more reliable and trustworthy when it performed a task competently compared to when it made errors, suggesting that task competency influences perceptions in robots \cite{salem}. In a similar vein, Carter et al. found that participants rated robots who did not interrupt during a task as more competent than robots that did interrupt the collaborative task \cite{carter}. Furthermore, Clair et al. employed verbal feedback in a human-robot collaboration task and found that ratings of team performance and the robot as a teammate were improved \cite{clair}. Although some work explored the interaction between verbal messages and task competency, there is a gap in the exploration of the effect of language fluency and task competency in HRI. Therefore, in this paper, we aim to examine the effects of varying task competency paired with varying language fluency on the perceptions of a social robot partner.

\section{RESEARCH QUESTIONS}

To examine the impact of varying language fluency and task competency on the perception of a social robot, we have designed four experimental conditions. In particular, we aim to investigate whether perceptions vary for a robot that is: (1) fluent and task competent; (2) fluent and task incompetent; (3) disfluent and task competent; or (4) disfluent and task incompetent. We refer to these four conditions as the varying robot conditions. Building on the literature, we want to investigate the impact of these varying robot conditions on the general perception of the robot (verbal competency, intelligence, reliability), the expectations being met, the robot as a teammate, and participants' willingness to work with the robot again. 

Therefore, we aimed to address the following four research questions: 

\begin{itemize}

\item \textbf{RQ1:} How do the perceptions (verbal competency, intelligence, and reliability) of the robot partner differ across the varying robot conditions? 
\item \textbf{RQ2:} How do the perceptions of the participants’ expectations being met differ across the varying robot conditions? 
\item \textbf{RQ3:} How do the perceptions of the robot as a teammate differ across the varying robot conditions? 
\item \textbf{RQ4:} How does the willingness to work with the robot differ across the varying robot conditions? 

\end{itemize}

\section{METHODS}

\subsection{What Am I? Study Design}

To address these research questions, we designed a human-robot interactive scenario loosely based on the guessing game, ``What Am I?''. This game consists of the interaction partner prompting the robot to guess the identity of an animal that the robot is assuming after hearing a few characteristics of that animal from the robot. The robot was positioned on a small table to face the human agent during the entire interaction. The participant was provided with a list of animals and their associated characteristics, which they could reference throughout the game. The participant was either assigned to a fluent robot, which spoke English at a native fluency level, or a disfluent robot, which spoke English at a non-native fluency level, for the entire duration of the game. We define fluency in detail in the following section. 

To initiate the game, the robot greets the player and asks if they are ready to begin the round. After the partner affirms, the robot states the characteristics of the animal whose identity it assumed and asks the participant, ``What am I?'' (fluent robot) or ``What I am?'' (disfluent robot). Then, the participant identifies the animal, and the robot either correctly confirms the accuracy of the identity or incorrectly rejects the participant’s selection and states the wrong animal for the characteristics it assumed. The robot then continues to the next round and repeats this process with a different animal identity until a total of three rounds are completed. At the conclusion of the third round, the robot states the game has ended and thanks the human partner for playing. Table~\ref{Sample Script} shows a sample script of the varying robot interactions.  

\subsection{Language Fluency and Task Competency Classification}

\noindent\textbf{Language Fluency:} There are many factors that can impact the spoken language fluency of a non-native speaker and the perception of their language fluency such as stutter, accent, hesitation, intonation, phonological processes, pronunciation rules, and ungrammatical words or sentence structures, etc. \cite{pinget, gurbuz, kormos, lennon}. 
For this study, we incorporated grammatical accuracy in a speech to represent the fluent robot partner. In order to accurately depict disfluency in the robot, we need to define speech disfluency. A possible taxonomy to understand speech disfluencies is by distinguishing speech disfluencies into two groups: disfluencies that stem from uncertainty and errors or error-type disfluencies (ETDs) \cite{bakti}. This taxonomy lists the principle measures of uncertainty-related speech disfluencies to include factors such as hesitations and repetition while error-type disfluencies encompass measures such as grammatical errors and contamination. In this work, we interpreted the root of speech disfluency to be ETDs. Specifically, we incorporated grammatical errors in a speech to represent the disfluent robot partner. We also chose this factor because grammatical errors can be made easily decipherable to participants in robot speech, and we wanted to avoid disfluency measures that could be interpreted as inherent to the robot. For example, stutter or hesitation could have been incorrectly attributed to or perceived as malfunctions or natural behavior of the robot. The types of grammatical errors implemented in the disfluent robot’s dialogue included inaccurate subject-verb agreement, incorrect singular/plural agreement, wrong pronoun usage (subjective vs  objective), incorrect word form, and lack of article and preposition usage. These errors were based on the most common grammatical errors made by non-native English speakers \cite{Setiyorini, alshujairi, kittiporn}. A sample script of the disfluent robot interaction is included in Table \ref{Sample Script}.

\begin{table}
  \caption{Sample Script for the Four Varying Robot Conditions.}
  \vspace{-0.05in}
  \label{Sample Script}
 \includegraphics[width=0.48\textwidth]{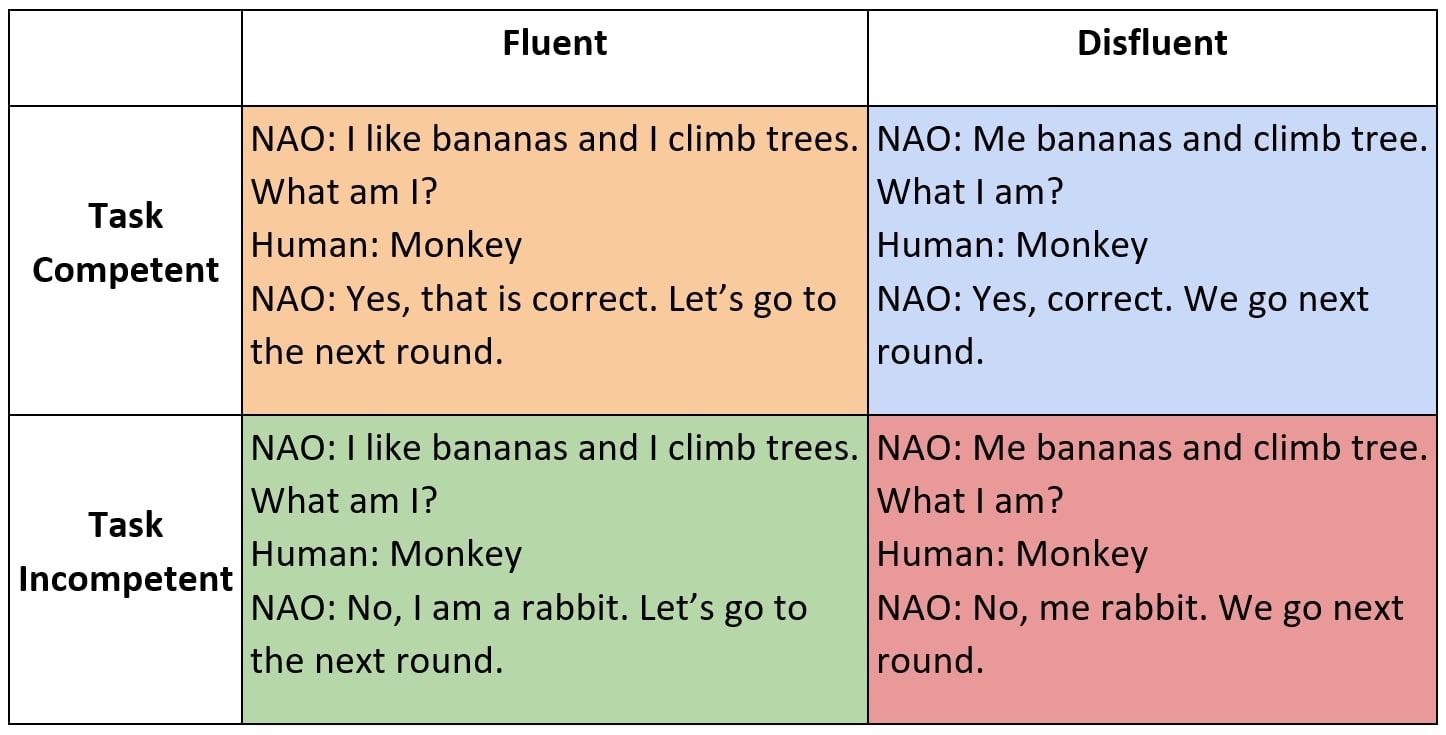}
 \vspace{-0.2in}
\end{table}

\noindent\textbf{Task competency:} Along with verbal language fluency, this study also looked at the perceptions of two types of task competencies by the robot: task competent (accurate task completion) and task incompetent (inaccurate task completion). The \textit{task competent condition} was demonstrated in the game by the robot correctly affirming the animal that the human player guessed to be the robot’s assumed identity. The \textit{task incompetent condition} was represented in the game by the robot incorrectly rejecting the animal that the human player guessed to be the robot’s assumed identity and stating a different incorrect animal as the answer. A sample script of the task competent and task incompetent robot interactions are included in Table \ref{Sample Script}.

\subsection{Participant Classification}

To reduce the participants' linguistic bias in the perception of the language fluency and task competency of the robot, we only chose to select monolingual participants to participate in this study. The monolingual categorization was based on participants' responses on a pre-task survey that collected biographical, demographic, and linguistic information and was completed upon arrival. The linguistic portion comprised two questions where the first asked how many languages the participant had verbal native fluency in, and the second followed up by asking to list the language(s) from the previous question. Native fluency was defined to the participants on the survey as being a native language that does not contain unnatural grammatical errors, pauses, stutters, repetitions, or self-corrections when speaking in the language. The participants of this study identified themselves as monolingual native English speakers. In order to be classified as a monolingual native speaker, the participant had to self-report their linguistic abilities and answer the first question with the number ``1'' and the second question with the language ``English''.

\subsection{The Robot}

In our study, the SoftBank Robotics’ NAO humanoid robot was used. To capture the robot’s nonverbal gestures, we utilized the expressive behavior modules in SoftBank’s Choregraphe Suite and for the robot’s verbal performance, we worked with Amazon Polly’s text-to-speech platform \cite{polly}. The decision to use Amazon Polly over Choegraphe’s text-to-speech option was due to Amazon Polly’s robust annunciation and timing capabilities. To mitigate potential bias with the perceptions of the robot’s gender, we use the gender-neutral ``ivy'' voice. Then, 18 sound clips of the fluent, disfluent, accurate task completion, and inaccurate task completion dialogue were added to the Choregraphe program and matched with some basic animation behaviors. In Choregraphe, we also utilized the speech recognition modules so that the robot could react accordingly to the participant's responses.

\section{EXPERIMENT}

\subsection{Procedure} \label{sec:procedure}

Participants reviewed a study information document for consent along with the task instructions and were asked to complete a brief demographic survey. Each participant was randomly assigned to one of four robot conditions. Next, the participants played three rounds of the ``What Am I?” game with the NAO robot. In each round, the robot would assume the identity of an animal and give a description of that animal to the participant, who would then have to guess the animal's identity based on the given description. After playing all three rounds, participants were asked to rate how much they agreed with the robot being verbally competent, intelligent, reliable, a good teammate, meeting their expectations, and willing to work with the robot again on a five-point Likert scale ranging from ``strongly disagree'' (1) to ``strongly agree” (5). The entire study took approximately 15 minutes. At the conclusion of the study, participants were debriefed and compensated.

\subsection{Participants}

A total of 60 adults participated in the study (70.0\% female ($n=42$), 30.0\% male ($n=18$)). The mean age of participants was 21 years ($SD=2.75$). All participants were required to be English speakers, available to participate in-person, at least 18 years of age or older, and monolingual. The participants consisted of undergraduate and graduate students, and high school and college graduates. During the study, participants also reported their experience with robots in general ($M=1.42$, $SD=0.740$). Participants were compensated with a \$5 gift card for participating in the study. The study was approved by the Institutional Review Board.

\subsection{Measures}

After the game, participants were asked to rate perceptions of the robot's verbal competency, intelligence, and reliability. We also had participants reflect on whether the NAO robot was a good teammate, rate their willingness to work with the robot again, and express whether the robot met their expectations on the Likert scale described in Sect.~\ref{sec:procedure}.

 \begin{figure*}[t]
    \centering
    \includegraphics[width=1\textwidth]{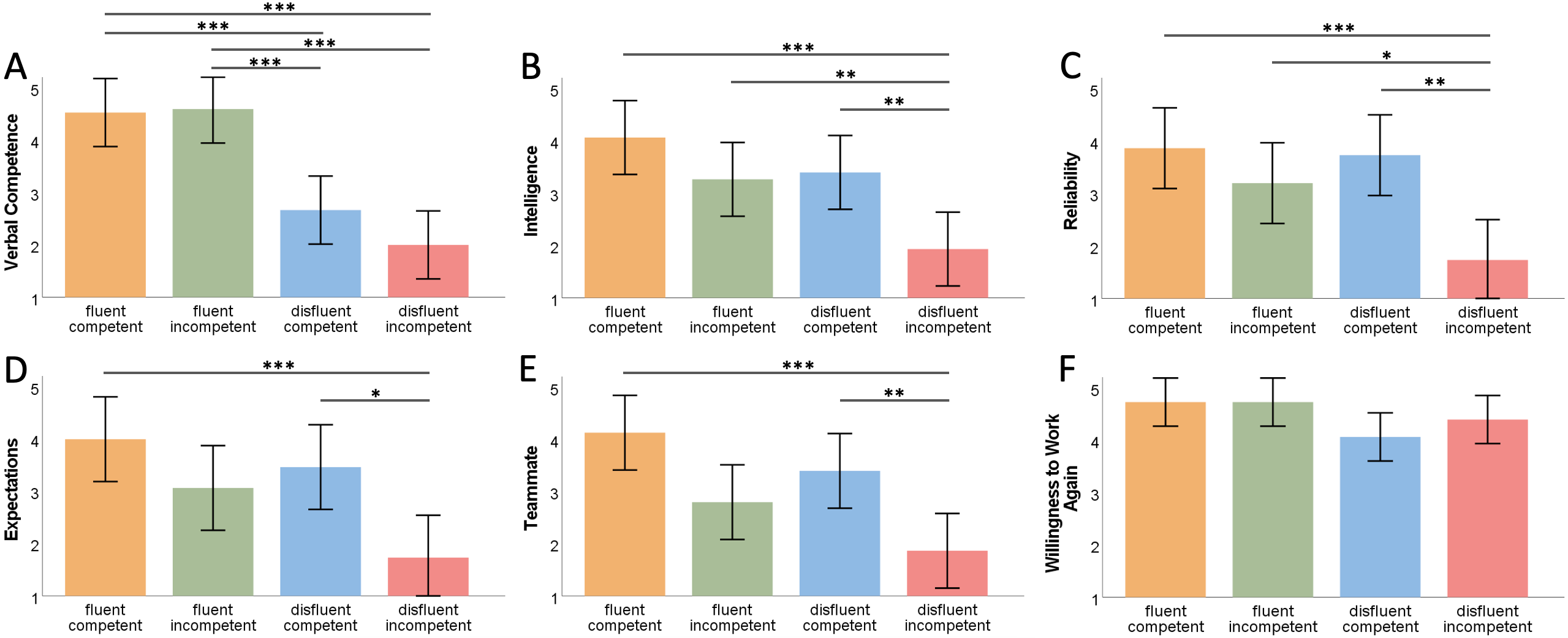}
    \caption{ Bar graphs showing participants' perceptions of verbal competence, intelligence, and reliability as well as their ratings of expectations being met, the robot as a teammate, and their willingness to work with the robot again across four varying robot conditions (the robot being fluent and task competent, fluent and task incompetent, disfluent and task competent, and disfluent and task incompetent). Error bars 95\% CI. The significant values are shown in * ($* = p < .05$, $** = p < .01$, $*** = p < .001$).}
    \label{Graph}
    \vspace{-0.2in}
\end{figure*}

\section{RESULTS AND DISCUSSION}

To address our research questions, we evaluated the effects of varying language fluency and task competency on the perceptions of a robot partner. A series of multivariate ANOVAs were conducted with four groups that make up the varying robot conditions as independent variables and six dependent variables (i.e., perceptions of verbal competence, intelligence, reliability, expectations being met, the robot as a teammate, and willingness to work with the robot again). For the analysis, an inspection of the data and Levene's test provided no strong evidence against the assumption of constant variance. Furthermore, we used Tukey’s Honestly Significant Difference (HSD) test for post-hoc comparisons. The multivariate analysis highlighted a statistically significant interaction effect between varying robot conditions on the combined perceptions of the robot partner, ($F(18, 144) = 4.999$, $p < .001$, Wilks' $\lambda = .255$, $\eta^2 = .366$). 

\subsection*{RQ1: Effects of Varying Robot Conditions on Perceptions}
 
 \subsubsection{Verbal Competence}
 Separate univariate ANOVAs on the dependent variables revealed significant effects of varying robot conditions on perceptions of verbal competence, ($F(3, 56) = 16.46$, $p < .001$, $\eta^2 = .469$). The results (Fig.~\ref{Graph}-A) suggest participants in the fluent-competent condition rated verbal competence ($M = 4.53$, $SD = 1.060$) higher than participants in the disfluent-competent group ($M = 2.67$, $SD = 1.397$), ($p < .001$). Participants in the fluent-competent condition rated verbal competence ($M = 4.53$, $SD = 1.060$) higher than participants in the disfluent-incompetent group ($M = 2.00$, $SD = 1.604$), ($p < .001$). Participants in the fluent-incompetent condition ($M = 4.60$, $SD = 0.828$) rated verbal competence higher than participants in the disfluent-competent group ($M = 2.67$, $SD = 1.397$), ($p < .001$). Participants in the fluent-incompetent condition rated verbal competence ($M = 4.60$, $SD = 0.828$) higher than participants in the disfluent-incompetent group ($M = 2.00$, $SD = 1.604$), ($p < .001$).

\noindent\textbf{Discussion:} The results indicate that participants found both the fluent robots (fluent-competent and fluent-incompetent) to be significantly more verbally competent than the disfluent robots (disfluent-competent and disfluent-incompetent). This was expected as the disfluent robots made grammatical errors, which may influence the perception of the verbal competency of those robots. The results may indicate that language fluency plays a greater role than task competency in the perception of verbal competency of a robot.  

\subsubsection{Intelligence}

Separate univariate ANOVAs on the dependent variables revealed significant effects of varying robot conditions on perceptions of intelligence ($F(3, 56) = 6.425$, $p < .001$, $\eta^2 = .256$). The results (Fig. \ref{Graph}-B) suggest participants in the fluent-competent condition rated intelligence ($M = 4.07$, $SD = 1.534$) higher than participants in the disfluent-incompetent condition ($M = 1.93$, $SD = 1.534$), ($p < .001$). Additionally, participants in the fluent-incompetent condition rated intelligence ($M = 3.27$, $SD = 1.163$) higher than participants in the disfluent-incompetent condition ($M = 1.93$, $SD = 1.534$), ($p = .047$). Participants in the disfluent-competent condition rated intelligence ($M = 3.40$, $SD = 1.183$) higher than participants in the disfluent-incompetent condition ($M = 1.93$, $SD = 1.534$), ($p = .024$).

\noindent\textbf{Discussion:} The results suggest that the participants perceived the fluent-competent, fluent-incompetent, and disfluent-competent robots to be more intelligent than the disfluent-incompetent robot. It was expected for the fluent-competent robot to be perceived as more intelligent than the disfluent-incompetent robot due to neither language fluency or task competency being compromised in the former condition. However, in the fluent-incompetent and disfluent-competent conditions, at least one of the variables (language fluency or task competency) was compromised, yet robots in both conditions were still perceived to be more intelligent than the disfluent-incompetent robot, which had both variables compromised. These results may suggest that both language fluency and task competency have an impact on the perceived intelligence of the robot.

\subsubsection{Reliability}

Separate univariate ANOVAs on the dependent variables revealed significant effects of varying robot conditions on perceptions of reliability ($F(3, 56) = 6.463$, $p < .001$, $\eta^2 = .257$). The results (Fig. \ref{Graph}-C) suggest participants in the fluent-competent condition rated reliability ($M = 3.87$, $SD = 1.807$) higher than participants in the disfluent-incompetent condition ($M = 1.73$, $SD = 1.534$), ($p = .001$). Additionally, participants in the fluent-incompetent condition rated reliability ($M = 3.20$, $SD = 1.207$) higher than participants in the disfluent-incompetent group ($M = 1.73$, $SD = 1.534$), ($p = .044$). Furthermore, participants in the disfluent-competent condition rated reliability ($M = 3.73$, $SD = 1.335$) higher than participants in the disfluent-incompetent condition ($M = 1.73$, $SD = 1.534$), ($p = .003$).

\noindent\textbf{Discussion:} The results indicate that participants in the fluent-competent, fluent-incompetent, and disfluent-competent conditions perceived the robot to be significantly more reliable than the participants in the disfluent-incompetent condition. Similarly to the intelligence results, it was expected for the fluent-competent robot to be perceived as more reliable than the disfluent-incompetent robot due to neither language fluency or task competency being compromised in the former condition. However, in the fluent-incompetent and disfluent-competent conditions, at least one of the variables (language fluency or task competency) was compromised, yet robots in both conditions were still perceived to be more reliable than the disfluent-incompetent robot, which had both variables compromised. The results may imply that both language fluency and task competency have an impact on the perceived reliability of the robot.

\subsection*{RQ2: Effects of Varying Robot Conditions on Participants’ Expectations Being Met }

Separate univariate ANOVAs on the dependent variables revealed significant differences between the varying robot conditions on ratings of expectations being met ($F(3, 56) = 5.716$, $p = .002$, $\eta^2 = .234$). Specifically, the results (Fig. \ref{Graph}-D) suggest participants in the fluent-competent condition rated that their expectations were met ($M = 4.00$, $SD = 1.732$) significantly higher than participants in the disfluent-incompetent condition ($M = 1.73$, $SD = 1.534$), ($p = .001$). Additionally, participants in the disfluent-competent condition rated that their expectations were met ($M = 3.47$, $SD = 1.356$) significantly higher than participants in the disfluent-incompetent conditions ($M = 1.73$, $SD = 1.534$), ($p = .019$).

\noindent\textbf{Discussion:} The results indicate that participants in the fluent-competent and disfluent-competent conditions perceived the robot to have met their expectations more than the participants in the disfluent-incompetent condition did. Additionally, there was no significant difference was found in the rating for the fluent-incompetent and disfluent-incompetent conditions. This result may suggest that task competency has a greater impact on the perception of the robot meeting expectations than language fluency.

\subsection*{RQ3: Effects of Varying Robot Conditions on Perception of Robot as a Teammate}

Separate univariate ANOVAs on the dependent variables revealed significant differences between the varying robot conditions on ratings of the robot as a teammate ($F(3, 56) = 7.188$, $p < .0005$, $\eta^2 = .278$). Specifically, the results (Fig. \ref{Graph}-E) suggest participants in the fluent-competent condition rated the robot as a teammate ($M = 4.13$, $SD = 1.356$) higher than participants in the disfluent-incompetent condition did ($M = 1.87$, $SD = 1.552$), ($p < .001$). Additionally, participants in the disfluent-competent condition ($M = 3.40$, $SD = 1.242$) rated the robot as a  teammate higher than participants in the disfluent-incompetent condition did ($M = 1.87$, $SD = 1.552$), ($p = .019$).

\noindent\textbf{Discussion:} Participants in the fluent-competent and disfluent-competent conditions rated the robot more highly for being a good teammate than participants in the disfluent-incompetent condition did. Additionally, there was no significant difference in the rating for the fluent-incompetent and disfluent-incompetent conditions. The results may indicate that task competency has a greater impact on the perception of the robot as a teammate than language fluency. 

\subsection*{RQ4: Effects of Varying Robot Conditions on Participants’ Willingness to Work with Robot Again}

There was no significance of the effects of varying robot conditions on participants' willingness to work with the robot again ($F(3, 56) = 1.927$, $p = .136$, $\eta^2 = .094$) (Fig. \ref{Graph}-F). 

\noindent \textbf{Discussion:} The results may suggest that neither language fluency nor task competency impacts whether participants would be willing to work with the robot again.

\section{OVERALL FINDINGS}


Overall, the fluent-competent robot was rated higher than the disfluent-incompetent robot in every perception category except willingness to work again. For \textbf{RQ1}, we observed a significant interaction between varying robot conditions on perceptions of the robot's verbal competence, intelligence, and reliability. One of the important findings is that language fluency plays a greater role than task competency on the impact of the robot's perceived verbal competence. Additionally, for the perceptions of intelligence and reliability, both language fluency and task competency have an impact on the participants' perceptions. 

For \textbf{RQ2} and \textbf{RQ3}, we explored the effect of varying robot conditions on ratings of the robot meeting expectations and the robot as a teammate. We discerned a significant interaction between varying robot conditions on perceptions of the robot meeting expectations and on the rating of the robot as a teammate. The results indicate that both the competent robots (fluent and disfluent) were perceived as better at meeting expectations and to be a good teammate than the disfluent-incompetent robot. Therefore, we can reason that task competency may be a factor of greater influence on the perceptions of the robot meeting expectations and the robot as a teammate. 

For \textbf{RQ4}, there were no significant differences between the effects of varying robot conditions on ratings of willingness to work with the robot again, which may indicate that neither language fluency nor task competency alone had a significant impact on the participant's rating of willingness to work with the robot again. We posit that this may be due to the tasks being relatively low stakes in their impact on participants. For example, the robot failing to perform the task correctly may not have been consequential enough in this context for the participant not to want to work with the robot again. There is a need for further exploration of the confounding factors that may impact the willingness of participants to work with the robot again.

\section{CONCLUSIONS}

In this paper, we presented a human-robot interaction study to investigate the effects of varying language fluency and task competency on the perception of robots. The results highlight that while language fluency may play a greater role than task competency in the perception of verbal competence, both language fluency and task competency contribute to the perception of intelligence and reliability of the robot. The results also indicate that task competency may play a greater role than language fluency in the perception of meeting expectations and being a good teammate. The findings of this work underscore the importance of language fluency and task competency in the context of social HRI and will enable the development of more intelligent robots in the future. In the future, we plan to examine how the perception of robots in the varying language fluency and task competency conditions from the perspective of multilingual individuals differs from that of the perspective of monolingual individuals. 









\bibliographystyle{IEEEtran}
\bibliography{IEEEabrv,sources}

\begin{thebibliography}{10}
\providecommand{\url}[1]{#1}
\csname url@rmstyle\endcsname
\providecommand{\newblock}{\relax}
\providecommand{\bibinfo}[2]{#2}
\providecommand\BIBentrySTDinterwordspacing{\spaceskip=0pt\relax}
\providecommand\BIBentryALTinterwordstretchfactor{4}
\providecommand\BIBentryALTinterwordspacing{\spaceskip=\fontdimen2\font plus
\BIBentryALTinterwordstretchfactor\fontdimen3\font minus \fontdimen4\font\relax}
\providecommand\BIBforeignlanguage[2]{{%
\expandafter\ifx\csname l@#1\endcsname\relax
\typeout{** WARNING: IEEEtran.bst: No hyphenation pattern has been}%
\typeout{** loaded for the language `#1'. Using the pattern for}%
\typeout{** the default language instead.}%
\else
\language=\csname l@#1\endcsname
\fi
#2}}

\bibitem{bahishti}
A.~A. Bahishti, ``Humanoid robots and human society,'' \emph{Advanced Journal of Social Science}, vol.~1, no.~1, p. 60–63, Nov. 2017.

\bibitem{yasar2020RAL}
M.~S. Yasar and T.~Iqbal, ``A scalable approach to predict multi-agent motion for human-robot collaboration,'' in \emph{IEEE Robotics and Automation Letters (RA-L)}, 2021.

\bibitem{islam2021multigat}
M.~M. Islam and T.~Iqbal, ``Multi-gat: A graphical attention-based hierarchical multimodal representation learning approach for human activity recognition,'' in \emph{IEEE Robotics and Automation Letters (RA-L)}, 2021.

\bibitem{unhelkar2020semi}
V.~V. Unhelkar, S.~Li, and J.~A. Shah, ``Semi-supervised learning of decision-making models for human-robot collaboration,'' in \emph{Conference on Robot Learning}, 2020.

\bibitem{maven}
M.~M. Islam, M.~S. Yasar, and T.~Iqbal, ``{MAVEN}: A memory augmented recurrent approach for multimodal fusion,'' in \emph{IEEE Transaction on Multimedia}, 2022.

\bibitem{caesar}
M.~M. Islam, R.~M. Mirzaiee, A.~Gladstone, H.~N. Green, and T.~Iqbal, ``{CAESAR}: A multimodal simulator for generating embodied relationship grounding dataset,'' in \emph{NeurIPS}, 2022.

\bibitem{kanda}
T.~Kanda, H.~Ishiguro, M.~Imai, and T.~Ono, ``Development and evaluation of interactive humanoid robots,'' \emph{Proceedings of the IEEE}, vol.~92, no.~11, pp. 1839--1850, 2004.

\bibitem{eqamx}
M.~M. Islam, A.~Gladstone, R.~Islam, and T.~Iqbal, ``Eqa-mx: Embodied question answering using multimodal expression,'' in \emph{The Twelfth International Conference on Learning Representations (ICLR)}, 2024.

\bibitem{vader}
M.~S. Yasar and T.~Iqbal, ``Vader: Vector-quantized generative adversarial network for motion prediction,'' in \emph{IEEE/RSJ International Conference on Intelligent Robots and Systems (IROS)}, 2023, pp. 3827--3834.

\bibitem{graaf}
M.~{de Graaf}, ``\BIBforeignlanguage{English}{Living with robots: investigating the user acceptance of social robots in domestic environments},'' Ph.D. dissertation, University of Twente, June 2015.

\bibitem{imprint}
M.~S. Yasar*, M.~M. Islam*, and T.~Iqbal, ``{IMPRINT}: Interactional dynamics-aware motion prediction in teams using multimodal context,'' in \emph{ACM Transactions on Human-Robot Interaction}, 2023.

\bibitem{patron}
M.~M. Islam, A.~Gladstone, and T.~Iqbal, ``{PATRON}: Perspective-aware multitask model for referring expression grounding using embodied multimodal cues,'' in \emph{Proceedings of the 37th AAAI Conference on Artificial Intelligence (AAAI)}, 2023.

\bibitem{yasar2022HRI}
M.~S. Yasar and T.~Iqbal, ``Robots that can anticipate and learn in human-robot teams,'' in \emph{2022 17th ACM/IEEE International Conference on Human-Robot Interaction (HRI)}, 2022.

\bibitem{yan}
Y.~Wang and J.~E. Young, ``2022 17th acm/ieee international conference on human-robot interaction (hri),'' 2014.

\bibitem{Iqbal2016T-RO}
T.~Iqbal, S.~Rack, and L.~D. Riek, ``Movement coordination in human-robot teams: A dynamical systems approach,'' \emph{IEEE Transactions on Robotics}, vol.~32, no.~4, pp. 909--919, 2016.

\bibitem{iqbal_2021}
T.~Iqbal and L.~D. Riek, ``Temporal anticipation and adaptation methods for fluent human-robot teaming,'' in \emph{2021 IEEE International Conference on Robotics and Automation (ICRA)}, 2021, pp. 3736--3743.

\bibitem{anzalone}
S.~M. Anzalone, S.~Boucenna, S.~Ivaldi, and M.~Chetouani, ``Evaluating the engagement with social robots,'' \emph{International Journal of Social Robotics}, vol.~7, pp. 465--478, 2015.

\bibitem{leite}
I.~Leite, C.~Martinho, and Paiva, ``Evaluating the engagement with social robots,'' \emph{International Journal of Social Robotics}, vol.~5, pp. 291--308, 2013.

\bibitem{friederike}
F.~Eyssel, D.~Kuchenbrandt, F.~Hegel, and L.~de~Ruiter, ``Activating elicited agent knowledge: How robot and user features shape the perception of social robots,'' in \emph{The 21st IEEE International Symposium on Robot and Human Interactive Communication}, 2012, pp. 851--857.

\bibitem{posetron}
M.~S. Yasar, M.~M. Islam, and T.~Iqbal, ``Posetron: Enabling close-proximity human-robot collaboration through multi-human motion prediction,'' in \emph{Proceedings of the ACM/IEEE International Conference on Human-Robot Interaction (HRI)}, 2024, pp. 830--839.

\bibitem{nikolaidis}
S.~Nikolaidis, M.~Kwon, J.~Forlizzi, and S.~Srinivasa, ``Planning with verbal communication for human-robot collaboration,'' vol.~7, no.~3, 2018.

\bibitem{vargas}
A.~Marin~Vargas, L.~Cominelli, F.~Dell’Orletta, and E.~P. Scilingo, ``Verbal communication in robotics: A study on salient terms, research fields and trends in the last decades based on a computational linguistic analysis,'' \emph{Frontiers in Computer Science}, vol.~2, 2021.

\bibitem{woo}
J.~Woo, J.~Botzheim, and N.~Kubota, ``Conversation system for natural communication with robot partner,'' in \emph{2014 10th France-Japan/ 8th Europe-Asia Congress on Mecatronics}, 2014, pp. 349--354.

\bibitem{haley_humor_aaai}
H.~N. Green, M.~M. Islam, S.~Ali, and T.~Iqbal, ``ispy a humorous robot: Evaluating the perceptions of humor types in a robot partner,'' in \emph{AAAI Spring Symposium on Putting AI in the Critical Loop: Assured Trust and Autonomy in Human-Machine Teams}, 2022.

\bibitem{zhao}
S.~Zhao, ``Humanoid social robots as a medium of communication,'' \emph{New Media \& Society}, vol.~8, no.~3, pp. 401--419, 2006.

\bibitem{haley_humor}
H.~N. Green, M.~M. Islam, S.~Ali, and T.~Iqbal, ``Who's laughing nao? examining perceptions of failure in a humorous robot partner,'' in \emph{ACM/IEEE International Conference on Human-Robot Interaction (HRI)}, 2022, p. 313–322.

\bibitem{breazeal}
C.~Breazeal, K.~Dautenhahn, and T.~Kanda, ``Social robotics,'' \emph{Springer handbook of robotics}, pp. 1935--1972, 2016.

\bibitem{grigore}
E.~C. Grigore, A.~Pereira, I.~Zhou, D.~Wang, and B.~Scassellati, ``Talk to me: Verbal communication improves perceptions of friendship and social presence in human-robot interaction,'' in \emph{16th International Conference Intelligent Virtual Agents}.\hskip 1em plus 0.5em minus 0.4em\relax Springer, 2016, pp. 51--63.

\bibitem{gluszek}
A.~Gluszek and J.~F. Dovidio, ``Speaking with a nonnative accent: Perceptions of bias, communication difficulties, and belonging in the united states.''

\bibitem{pavlenko}
A.~Pavlenko and A.~Blackledge, \emph{Negotiation of identities in multilingual contexts}.\hskip 1em plus 0.5em minus 0.4em\relax Multilingual Matters, 2004, vol.~45.

\bibitem{kang}
O.~Kang, D.~Rubin, and S.~Lindemann, ``Mitigating us undergraduates’ attitudes toward international teaching assistants,'' \emph{Tesol Quarterly}, vol.~49, no.~4, pp. 681--706, 2015.

\bibitem{van}
S.~van Waveren, E.~J. Carter, and I.~Leite, ``Take one for the team: The effects of error severity in collaborative tasks with social robots,'' in \emph{Proceedings of the 19th ACM International Conference on Intelligent Virtual Agents}, 2019, pp. 151--158.

\bibitem{walker}
N.~Walker, C.~Mavrogiannis, S.~Srinivasa, and M.~Cakmak, ``Influencing behavioral attributions to robot motion during task execution,'' in \emph{Proceedings of the 5th Conference on Robot Learning}, vol. 164, 08--11 Nov 2022, pp. 169--179.

\bibitem{haring}
M.~H\"{a}ring, D.~Kuchenbrandt, and E.~Andr\'{e}, ``Would you like to play with me? how robots' group membership and task features influence human-robot interaction,'' in \emph{Proceedings of the 2014 ACM/IEEE International Conference on Human-Robot Interaction}, 2014, p. 9–16.

\bibitem{kim}
B.~Kim, K.~S. Haring, H.~J. Schellin, T.~N. Oberley, K.~M. Patterson, E.~Phillips, E.~J. de~Visser, and C.~C. Tossell, ``How early task success affects attitudes toward social robots,'' in \emph{Companion of the 2020 ACM/IEEE International Conference on Human-Robot Interaction}, 2020, p. 287–289.

\bibitem{jung}
M.~F. Jung, J.~J. Lee, N.~DePalma, S.~O. Adalgeirsson, P.~J. Hinds, and C.~Breazeal, ``Engaging robots: Easing complex human-robot teamwork using backchanneling,'' in \emph{Proceedings of the 2013 Conference on Computer Supported Cooperative Work}, 2013, p. 1555–1566.

\bibitem{waveren}
S.~van Waveren, E.~J. Carter, and I.~Leite, ``Take one for the team: The effects of error severity in collaborative tasks with social robots,'' in \emph{Proceedings of the 19th ACM International Conference on Intelligent Virtual Agents}, 2019, p. 151–158.

\bibitem{carter}
E.~J. Carter, L.~M. Hiatt, and S.~Rosenthal, ``You're delaying my task?! the impact of task order and motive on perceptions of a robot,'' in \emph{2022 17th ACM/IEEE International Conference on Human-Robot Interaction (HRI)}, 2022, pp. 304--312.

\bibitem{salem}
M.~Salem, G.~Lakatos, F.~Amirabdollahian, and K.~Dautenhahn, ``Would you trust a (faulty) robot? effects of error, task type and personality on human-robot cooperation and trust,'' 2015, p. 141–148.

\bibitem{lennon}
P.~Lennon, ``Investigating fluency in efl: A quantitative approach,'' \emph{Language Learning}, vol.~40, no.~3, pp. 387--417, 1990.

\bibitem{vanpatten}
B.~VanPatten and J.~Williams, ``Skill acquisition theory robert dekeyser,'' in \emph{Theories in Second Language Acquisition}.\hskip 1em plus 0.5em minus 0.4em\relax Routledge, 2014, pp. 106--124.

\bibitem{kormos}
J.~Kormos and M.~D{\'e}nes, ``Exploring measures and perceptions of fluency in the speech of second language learners,'' \emph{System}, vol.~32, no.~2, pp. 145--164, 2004.

\bibitem{bosker}
H.~R. Bosker, A.-F. Pinget, H.~Quené, T.~Sanders, and N.~H. de~Jong, ``What makes speech sound fluent? the contributions of pauses, speed and repairs,'' \emph{Language Testing}, vol.~30, no.~2, pp. 159--175, 2013.

\bibitem{white}
M.~J. White and Y.~Li, ``Second-language fluency and person perception in china and the united states,'' \emph{Journal of Language and Social Psychology}, vol.~10, no.~2, pp. 99--113, 1991.

\bibitem{bongaerts}
T.~Bongaerts, C.~van Summeren, B.~Planken, and E.~Schils, ``Age and ultimate attainment in the pronunciation of a foreign language,'' \emph{Studies in second language acquisition}, pp. 447--465, 1997.

\bibitem{hagi}
A.~Hagi-Mohamed, ``Perceptions of nonnative english-speaking graduate teaching assistants: Identity issues, successes, and challenges in the field of tesl/tesol,'' \emph{Culminating Projects in English}, 2018.

\bibitem{barona}
D.~Barona, ``Native and non-native speakers’ perceptions of non-native accents,'' \emph{LL Journal}, vol.~3, no.~2, pp. 1--18, 2008.

\bibitem{munro}
M.~J. Munro and T.~M. Derwing, ``The functional load principle in esl pronunciation instruction: An exploratory study,'' \emph{System}, vol.~34, no.~4, pp. 520--531, 2006.

\bibitem{rubin}
D.~L. Rubin and K.~A. Smith, ``Effects of accent, ethnicity, and lecture topic on undergraduates' perceptions of nonnative english-speaking teaching assistants,'' \emph{International journal of intercultural relations}, vol.~14, no.~3, pp. 337--353, 1990.

\bibitem{todd}
R.~W. Todd and P.~Pojanapunya, ``Implicit attitudes towards native and non-native speaker teachers,'' \emph{System}, vol.~37, no.~1, pp. 23--33, 2009.

\bibitem{james}
A.~James and J.~Leather, \emph{Second-language speech: structure and process}.\hskip 1em plus 0.5em minus 0.4em\relax Walter de Gruyter, 2011, vol.~13.

\bibitem{groff}
C.~Groff, W.~Zwaanswijk, A.~Wilson, and N.~Saab, ``Language diversity as resource or as problem? educator discourses and language policy at high schools in the netherlands,'' \emph{International Multilingual Research Journal}, pp. 1--19, 2023.

\bibitem{jackson}
R.~Jackson and T.~Williams, ``On perceived social and moral agency in natural language capable robots,'' 03 2020.

\bibitem{scheutz_cantrell}
M.~Scheutz, R.~Cantrell, and P.~Schermerhorn, ``Toward humanlike task-based dialogue processing for human robot interaction,'' \emph{AI Magazine}, vol.~32, no.~4, pp. 77--84, 2011.

\bibitem{bisk}
Y.~Bisk, D.~Yuret, and D.~Marcu, ``Natural language communication with robots,'' in \emph{Proceedings of the 2016 Conference of the North {A}merican Chapter of the Association for Computational Linguistics: Human Language Technologies}, 2016, pp. 751--761.

\bibitem{wang}
N.~Wang, D.~V. Pynadath, S.~G. Hill, and A.~P. Ground, ``Building trust in a human-robot team with automatically generated explanations,'' in \emph{Proceedings of the interservice/industry training, simulation and education conference (I/ITSEC)}, vol. 15315, 2015, pp. 1--12.

\bibitem{wenger}
E.~Wenger, M.~Bronckers, C.~Cianfarani, J.~Cryan, A.~Sha, H.~Zheng, and B.~Y. Zhao, ``" hello, it's me": Deep learning-based speech synthesis attacks in the real world,'' in \emph{Proceedings of the 2021 ACM SIGSAC Conference on Computer and Communications Security}, 2021, pp. 235--251.

\bibitem{clair}
A.~St.~Clair and M.~Mataric, ``How robot verbal feedback can improve team performance in human-robot task collaborations,'' in \emph{ACM/IEEE International Conference on Human-Robot Interaction}, 2015, p. 213–220.

\bibitem{pinget}
A.-F. Pinget, H.~R. Bosker, H.~Quen{\'e}, and N.~H. De~Jong, ``Native speakers’ perceptions of fluency and accent in l2 speech,'' \emph{Language Testing}, vol.~31, no.~3, pp. 349--365, 2014.

\bibitem{gurbuz}
N.~G{\"u}rb{\"u}z, ``Understanding fluency and disfluency in non-native speakers' conversational english,'' \emph{Educational Sciences: Theory \& Practice}, 2017.

\bibitem{bakti}
M.~Bakti, ``Speech disfluencies in simultaneous interpretation,'' \emph{De Crom, Dies (ed.)}, 2009.

\bibitem{Setiyorini}
T.~J. Setiyorini, P.~Dewi, and E.~S. Masykuri, ``The grammatical error analysis found in students’ composition,'' \emph{Journal Lensa}, vol.~10, no.~2, 2020.

\bibitem{alshujairi}
Y.~B. Al-Shujairi and H.~Tan, ``Grammar errors in the writing of iraqi english language learners,'' \emph{International Journal of Education \& Literacy Studies}, vol.~5, no.~4, 2017.

\bibitem{kittiporn}
K.~Nonkukhetkhong, ``Grammatical errors analysis of the first year english major students, udon thani rajabhat university,'' in \emph{ACLL2013 Conference Proceedings}.\hskip 1em plus 0.5em minus 0.4em\relax IAFOR, 2013.

\bibitem{polly}
{Amazon Web Services, Inc}, ``Amazon polly,'' \url{https://aws.amazon.com/polly/}, 2021, retrieved: June, 2021.

\end{thebibliography}

\end{document}